\makeatletter
\newcommand{\dontusepackage}[2][]{%
  \@namedef{ver@#2.sty}{9999/12/31}%
  \@namedef{opt@#2.sty}{#1}}
\makeatother
\dontusepackage{subfigure}

\documentclass[]{article}
\usepackage{bbm}
\usepackage{lmodern}
\usepackage{amssymb,amsmath}
\usepackage{ifxetex,ifluatex}
\usepackage[usenames,dvipsnames]{color}
\usepackage{fixltx2e} % provides \textsubscript
\ifnum 0\ifxetex 1\fi\ifluatex 1\fi=0 % if pdftex
  \usepackage[T1]{fontenc}
  \usepackage[utf8]{inputenc}
\else % if luatex or xelatex
  \ifxetex
    \usepackage{mathspec}
    \usepackage{xltxtra,xunicode}
  \else
    \usepackage{fontspec}
  \fi
  \defaultfontfeatures{Mapping=tex-text,Scale=MatchLowercase}
  
\fi
\IfFileExists{upquote.sty}{\usepackage{upquote}}{}
\IfFileExists{microtype.sty}{%
\usepackage{microtype}
\UseMicrotypeSet[protrusion]{basicmath} % disable protrusion for tt fonts
}{}
\usepackage[margin=1.0in,bottom=1.5in]{geometry}
\usepackage[square,numbers,sort&compress]{natbib}
\usepackage{listings}
\usepackage{graphicx}
\makeatletter
\def\maxwidth{\ifdim\Gin@nat@width>\linewidth\linewidth\else\Gin@nat@width\fi}
\def\maxheight{\ifdim\Gin@nat@height>\textheight\textheight\else\Gin@nat@height\fi}
\makeatother
\setkeys{Gin}{width=\maxwidth,height=\maxheight,keepaspectratio}
\usepackage{caption}
\usepackage{float}
\usepackage{subfig}
\ifxetex
  \usepackage[setpagesize=false, % page size defined by xetex
              unicode=false, % unicode breaks when used with xetex
              xetex]{hyperref}
\else
  \usepackage[unicode=true]{hyperref}
\fi
\hypersetup{breaklinks=true,
            bookmarks=true,
            pdfauthor={},
            pdftitle={Practical Bayesian experimental design with conditional normalizing flows},
            colorlinks=true,
            citecolor=black,
            urlcolor=blue,
            linkcolor=black,
            pdfborder={0 0 0}}
\usepackage[preprint]{neurips_2019}
\usepackage{url}
\usepackage{booktabs}
\usepackage{amsfonts}
\usepackage{nicefrac}
\usepackage{microtype}

\makeatletter
\let\@oldfnsymbol\@fnsymbol
\renewcommand{\@fnsymbol}[1]{\@oldfnsymbol{0}}
\makeatother

\title{BEACON: Bayesian Experimental design Acceleration with Conditional Normalizing flows – a case study in optimal monitor well placement for CO$_2$ sequestration}
\author{
    Rafael Orozco\textsuperscript{*}\thanks{* Correspondence: \href{mailto:rorozco@gatech.edu}{rorozco@gatech.edu}},
    Abhinav Gahlot, and
    Felix J. Herrmann\\
    Georgia Institute of Technology
}
\date{}

\begin{document}
\maketitle
\begin{abstract}
CO$_2$ sequestration is a crucial engineering solution for mitigating climate change. However, the uncertain nature of reservoir properties, necessitates rigorous monitoring of CO$_2$ plumes to prevent risks such as leakage, induced seismicity, or breaching licensed boundaries. To address this, project managers use borehole wells for direct CO$_2$ and pressure monitoring at specific locations. Given the high costs associated with drilling, it is crucial to strategically place a limited number of wells to ensure maximally effective monitoring within budgetary constraints. Our approach for selecting well locations integrates fluid-flow solvers for forecasting plume trajectories with generative neural networks for plume inference uncertainty. Our methodology is extensible to three-dimensional domains and is developed within a Bayesian framework for optimal experimental design, ensuring scalability and mathematical optimality. We use a realistic case study to verify these claims by demonstrating our method's application in a large scale domains and optimal performance as compared to baseline well placement.
\end{abstract}

\section{Methodology}\label{methodology}
We guide the optimal placement of wells through the Bayesian principle of drilling at locations that provide most information gain as compared to our prior knowledge of the plume $\mathbf{x}_k$ at time step $k$. This principle is expressed numerically by the Expected Information Gain (EIG) and defined as the distributional difference (measured by the Kullback-Leibler divergence) between the prior and the posterior conditioned on data $\mathbf{y}_k$ acquired from the wells placed with design $\mathbf{M}$:

\begin{align}
EIG(\mathbf{M}) =& \mathbb{E}_{p(\mathbf{y}_k|\mathbf{M})}\,\,\left[ D_{KL} (p(\mathbf{x}_k|\mathbf{y}_k)\, || \, p(\mathbf{x}_k)) \right]. 
 \end{align} 
 
 Due to the maximum-likelihood training objective of normalizing flows \cite{dinh2014nice}, they directly measure the Kullback-Leibler divergence thus are naturally equipped to solve the optimal well-placement problem by means of increasing EIG \cite{foster2020unified,orozco2024probabilistic}. Moreover, we interpret optimal locations not as fixed points but as probabilistic densities that suggest areas with a higher likelihood of drilling similar to \cite{wu2021automated}. We combine training of a normalizing flow $f_\theta$ and optimization of probability density of well locations into the joint optimization:

 \begin{equation} \label{eq:opt-joint}
    \hat \theta,  \mathbf{\hat w} = \underset{\theta,\, \mathbf{w}}{\operatorname{arg max}} \,  \frac{1}{N} \sum_{i=1}^{N}  \left( -\frac{1}{2}\lVert  f_{\theta}(\mathbf{x}_k^{(i)};  \mathbf{M}(\mathbf{w}) \odot \mathbf{y}_k^{(i)}) \rVert_{2}^2 + \log \left|  \det{  \mathbf{J}_{f_{\theta}}} \right| \right),
\end{equation} 

 where we sample from the well density while enforcing well budget $s$ using the following expressions:

 \begin{align} \label{eq:binarize}
 \mathbf{M}(\mathbf{w}) := \mathbbm{1}_{\text{s}\frac{\mathbf{w}}{\mathbf{\overline  w} }  < \mathbf{u} } \\ 
 \text{where} \, \, \, \mathbf{u} \sim U(0,1).
\end{align}
 
 To create the training examples $\{\mathbf{x}_k^{(i)},\mathbf{y}_k^{(i)}\}_{i=0}^{i=N}$, we use fluid physics solvers \cite{louboutin2022accelerating,yin2023slimgroup} to propagate the knowledge of the plume at time step $k-1$ (expressed as a set of prior samples $\{\mathbf{x}_{k-1}^{(i)}\}_{i=0}^{i=N}$) to forecasts at time step $k$ that account for the inherent uncertainties in estimated subsurface permeability. We then add corruption noise $\varepsilon$ to the plume simulations $\mathbf{y}_k^{(i)} = \mathbf{x}_k^{(i)} + \varepsilon$ such that they represent the fully sampled plume observation over which we can sample wells by the binary mask operator $\mathbf{M}(\mathbf{w})$.  After training, our algorithm has two outputs: a probabilistic density for the optimal well placement $\mathbf{\hat w}$ and an amortized generative network $f_{\hat \theta}$ capable of generating samples of the unknown plume given the field observations. We also provide imaged seismic as inputs to the network but we do not perform their optimal placement. Future work, will explore optimal seismic imaging acquisition for probabilistic seismic imaging workflows as in \cite{yin2023wise}.

 We denote our general algorithm \textbf{BEACON}: \textbf{B}ayesian \textbf{E}xperimental design \textbf{A}cceleration with \textbf{CO}nditional \textbf{N}ormalizing flows. For more details, refer to our previous work where we introduced this algorithm in the context of medical imaging \cite{orozco2024probabilistic}. In this case study, we test BEACON by placing it inside an uncertainty-aware Digital Twin \cite{herrmann2023president,gahlot2023inference}.
 
 \begin{figure}
\centering
%\captionsetup[subfigure]{labelformat=empty}
{\includegraphics[width=0.99\hsize]{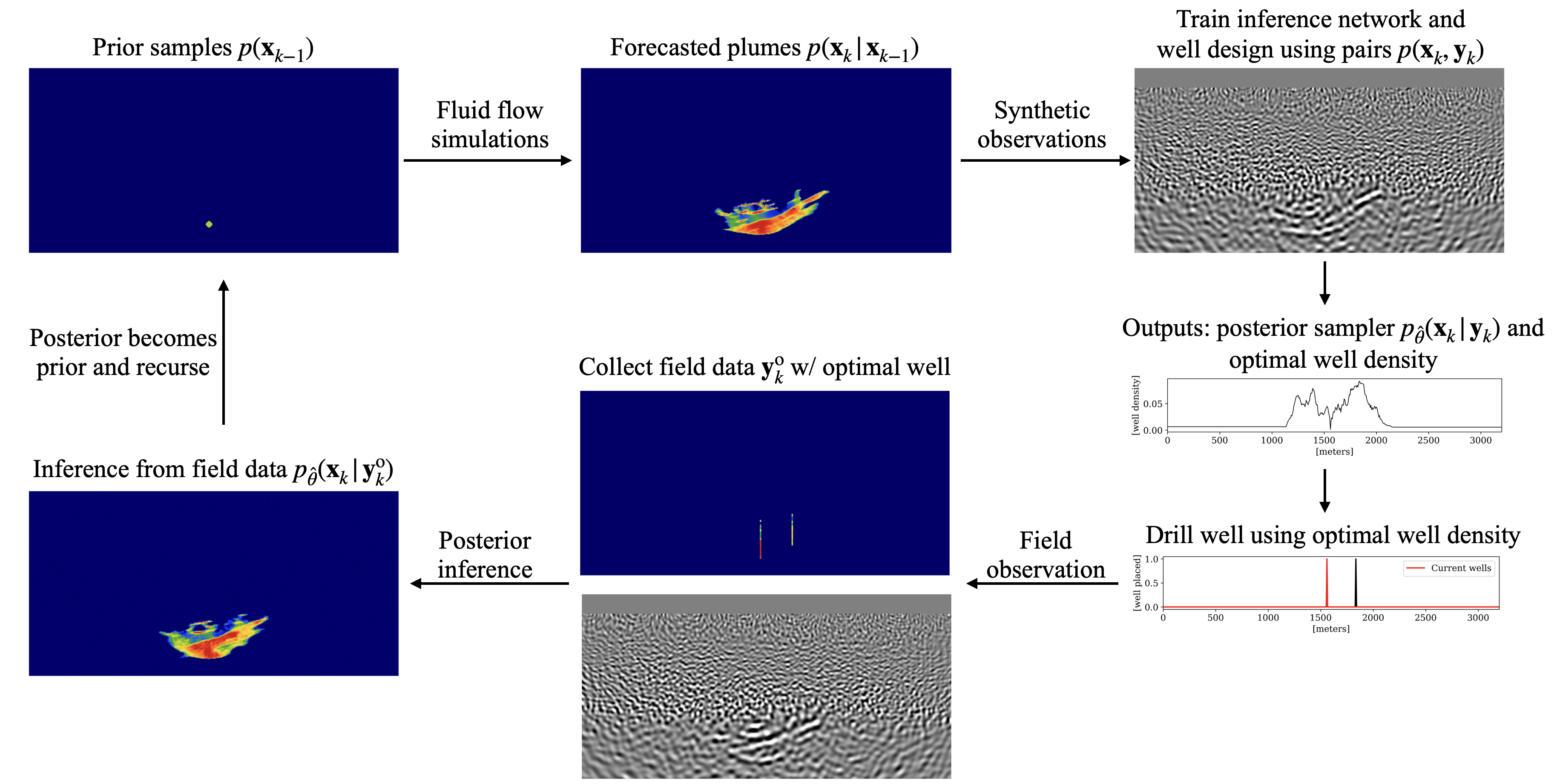}}
\caption{BEACON workflow integrated to digital twin for CO$_2$ sequestration monitoring. The sequence starts with prior samples of the plumes that are passed through the fluid simulator to generate plume forecasts. These forecasts are used to define the training observations of both well observations and seismic images. After training the normalizing flow and finding the optimal well location density, we simulate a field observation $\mathbf{y}_k^{o}$ from the ground truth plume and sample from the amortized posterior. The posterior samples become the prior samples of the next iteration and the sequence continues.}\label{fig-workflow}
\end{figure}

\section{Results and Conclusions}\label{results}
\begin{figure}
\centering
%\captionsetup[subfigure]{labelformat=empty}
{\includegraphics[width=0.99\hsize]{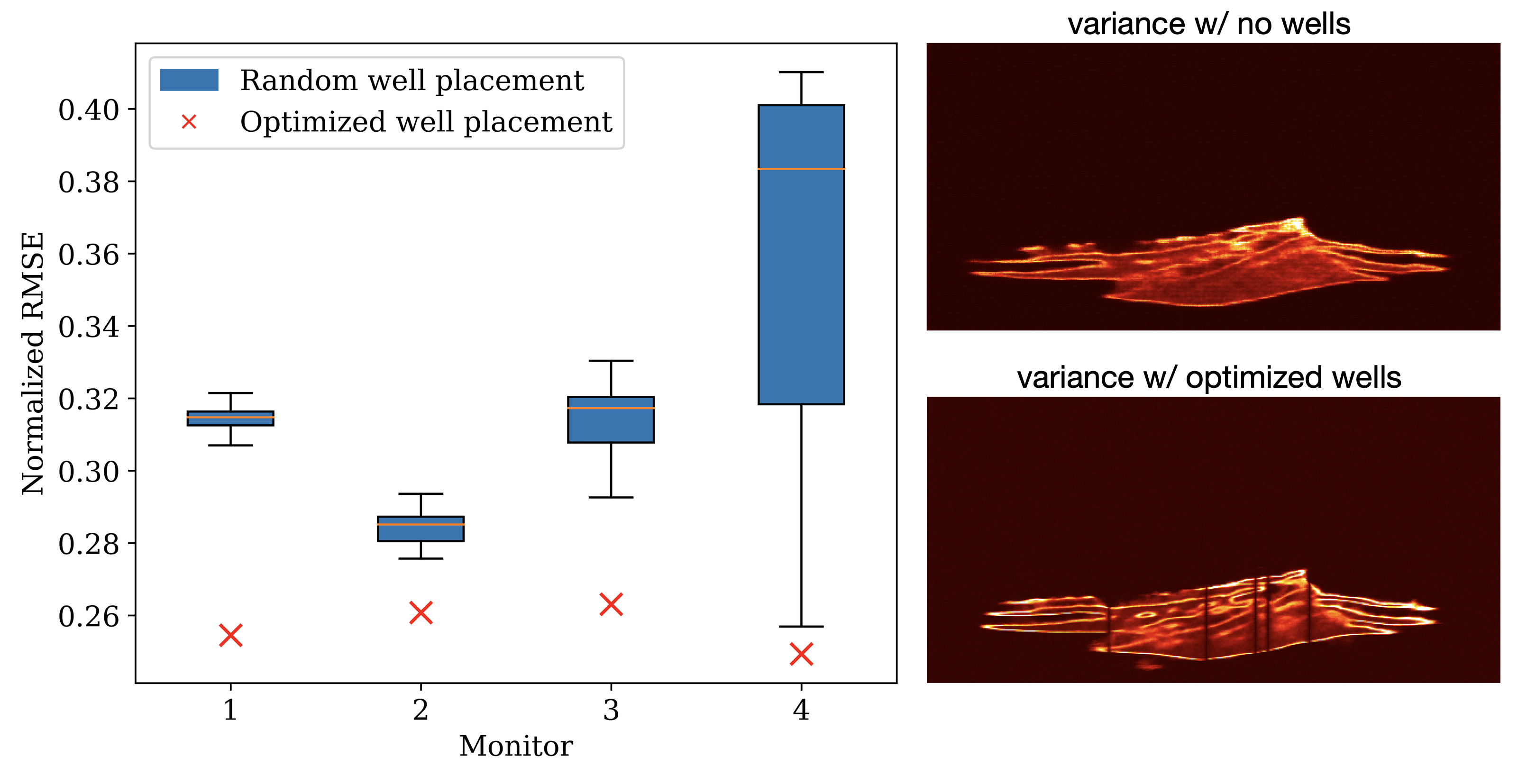}}
\caption{BEACON compared with random well placement baseline. }\label{fig-results}
\end{figure}

We design our algorithm to evolve alongside the CO$_2$ sequestration project's life cycle. This adaptability is achieved through integration into a Digital Twin workflow, which enables continuous refinement of well placement strategies based on incoming field observations. Figure \ref{fig-workflow} illustrates a single iteration, starting with prior samples of the plume and culminating in the generation of posterior samples post-observation at newly drilled optimally located wells. This process is iterative, with posterior distributions from one cycle becoming the priors for the subsequent cycle and one optimal well being drilled at each iteration for a total of $4$ iterations. In our synthetic case study, we employed realistic samples for permeability models derived from the Compass dataset \cite{jones2012building}. We juxtaposed our optimized well placements against a set of randomly selected well locations. We measure error rates by comparing the inferred posterior mean against the known synthetic ground truth and calculating the root mean squared error. The results, depicted in Figure \ref{fig-results}, show the benefits of our approach: not only is the uncertainty in plume inferences reduced but consistently lower error rates in comparison to the baseline are also obtained. This highlights the value of our algorithm BEACON for well placement to enhance the monitoring and management of CO$_2$ sequestration projects.

\section{Acknowledgement}\label{acknowledgement}

This research was carried out with the support of Georgia Research Alliance and partners of the ML4Seismic Center. 

\bibliography{paper}

\begin{thebibliography}{10}
\providecommand{\natexlab}[1]{#1}
\providecommand{\url}[1]{\texttt{#1}}
\expandafter\ifx\csname urlstyle\endcsname\relax
  \providecommand{\doi}[1]{doi: #1}\else
  \providecommand{\doi}{doi: \begingroup \urlstyle{rm}\Url}\fi

\bibitem[Dinh et~al.(2014)Dinh, Krueger, and Bengio]{dinh2014nice}
Laurent Dinh, David Krueger, and Yoshua Bengio.
\newblock Nice: Non-linear independent components estimation.
\newblock \emph{arXiv preprint arXiv:1410.8516}, 2014.

\bibitem[Foster et~al.(2020)Foster, Jankowiak, O’Meara, Teh, and Rainforth]{foster2020unified}
Adam Foster, Martin Jankowiak, Matthew O’Meara, Yee~Whye Teh, and Tom Rainforth.
\newblock A unified stochastic gradient approach to designing bayesian-optimal experiments.
\newblock In \emph{International Conference on Artificial Intelligence and Statistics}, pages 2959--2969. PMLR, 2020.

\bibitem[Orozco et~al.(2024)Orozco, Herrmann, and Chen]{orozco2024probabilistic}
Rafael Orozco, Felix~J Herrmann, and Peng Chen.
\newblock Probabilistic bayesian optimal experimental design using conditional normalizing flows.
\newblock \emph{arXiv preprint arXiv:2402.18337}, 2024.

\bibitem[Wu et~al.(2021)Wu, Verschuur, and Blacqui{\`e}re]{wu2021automated}
Sixue Wu, Dirk~J Verschuur, and Gerrit Blacqui{\`e}re.
\newblock Automated seismic acquisition geometry design for optimized illumination at the target: A linearized approach.
\newblock \emph{IEEE Transactions on Geoscience and Remote Sensing}, 60:\penalty0 1--13, 2021.

\bibitem[Louboutin et~al.(2022)Louboutin, Witte, Siahkoohi, Rizzuti, Yin, Orozco, and Herrmann]{louboutin2022accelerating}
Mathias Louboutin, Philipp Witte, Ali Siahkoohi, Gabrio Rizzuti, Ziyi Yin, Rafael Orozco, and Felix~J Herrmann.
\newblock Accelerating innovation with software abstractions for scalable computational geophysics.
\newblock In \emph{Second International Meeting for Applied Geoscience \& Energy}, pages 1482--1486. Society of Exploration Geophysicists and American Association of Petroleum~…, 2022.

\bibitem[Yin et~al.(2023{\natexlab{a}})Yin, Bruer, and Louboutin]{yin2023slimgroup}
Ziyi Yin, Grant Bruer, and Mathias Louboutin.
\newblock Slimgroup/jutuldarcyrules. jl: V0. 2.5 (version v0. 2.5). zenodo, 2023{\natexlab{a}}.

\bibitem[Yin et~al.(2023{\natexlab{b}})Yin, Orozco, Louboutin, and Herrmann]{yin2023wise}
Ziyi Yin, Rafael Orozco, Mathias Louboutin, and Felix~J Herrmann.
\newblock Wise: full-waveform variational inference via subsurface extensions.
\newblock \emph{arXiv preprint arXiv:2401.06230}, 2023{\natexlab{b}}.

\bibitem[Herrmann(2023)]{herrmann2023president}
Felix~J Herrmann.
\newblock President's page: Digital twins in the era of generative ai.
\newblock \emph{The Leading Edge}, 42\penalty0 (11):\penalty0 730--732, 2023.

\bibitem[Gahlot et~al.(2023)Gahlot, Erdinc, Orozco, Yin, and Herrmann]{gahlot2023inference}
Abhinav~Prakash Gahlot, Huseyin~Tuna Erdinc, Rafael Orozco, Ziyi Yin, and Felix~J Herrmann.
\newblock Inference of co2 flow patterns--a feasibility study.
\newblock \emph{arXiv preprint arXiv:2311.00290}, 2023.

\bibitem[Jones et~al.(2012)Jones, Edgar, Selvage, and Crook]{jones2012building}
CE~Jones, JA~Edgar, JI~Selvage, and H~Crook.
\newblock Building complex synthetic models to evaluate acquisition geometries and velocity inversion technologies.
\newblock In \emph{74th EAGE Conference and Exhibition incorporating EUROPEC 2012}, pages cp--293. European Association of Geoscientists \& Engineers, 2012.

\end{thebibliography}

\end{document}